\documentclass[a4paper]{article}

\usepackage[english]{babel}
\usepackage[utf8x]{inputenc}
\usepackage[T1]{fontenc}

\usepackage[a4paper,top=3cm,bottom=2cm,left=3cm,right=3cm,marginparwidth=1.75cm]{geometry}

\usepackage{amsmath}
\usepackage{graphicx}
\usepackage[colorinlistoftodos]{todonotes}
\usepackage[colorlinks=true, allcolors=blue]{hyperref}
\usepackage{comment}
\usepackage{parskip}
\usepackage{tabularx}
\usepackage{adjustbox}

\title{Patient similarity: methods and applications}
\author{Leyu Dai,He Zhu,Dianbo Liu\\
Computer science,The George Washington University\\
The Hong Kong Polytechnic University\\
Department of Biomedical Informatics, Harvard University\\
Computer Science and artificial Intelligence Laboratory, MIT\\
Corresponds to: dianbo@mit.edu
}

\begin{document}
\maketitle

\begin{abstract}
Patient similarity analysis is important in health care applications. It takes patient information such as their electronic medical records and genetic data as input and computes the pairwise similarity between patients. Procedures of typical a patient similarity study can be divided into several steps including data integration, similarity measurement, and neighborhood identification. And according to an analysis of patient similarity, doctors can easily find the most suitable treatments. There are many methods to analyze the similarity such as cluster analysis. And during machine learning become more and more popular, Using neural networks such as CNN is a new hot topic. This review summarizes representative methods used in each step and discusses applications of patient similarity networks especially in the context of precision medicine. 
\end{abstract}

\section{Introduction}

With the development of digital healthcare systems and high throughput laboratory technologies, a variety of patient-specific data such as diagnoses, treatment records, biochemical tests, genetic information become electronically available\cite{Kohane2015TenMedicine}. On the other hand, lacking qualified and specialized physicians has become a problem in many parts of the world. Ability to automatically build a patient similarity network without incurring additional efforts physicians can improve the efficiency of our health care systems and benefit both patients and hospitals. One field that patient similarity would contribute to is precision medicine. According to the Precision Medicine Initiative, precision medicine is "an emerging approach for disease treatment and prevention that takes into account individual variability in genes, environment, and lifestyle for each person"\cite{garrido2018proposal}. Putting it differently, precision medicine is the personalization of health care. 

There is general public enthusiasm around harnessing the personalized information in big data collected from electronic medical records, human genome sequencing, environmental factors, and many other aspects \cite{MITtechnologyreview2014Data-DrivenCare}. With the right patient similarity network built from large scale data in place, physicians can retrieve a cohort of similar patients for a target patient based on the case of study, make medical comparisons and, thereafter, make personalized treatment plan effectively.  One goal of precision medicine is to build quantitative models that could assist clinical decision making\cite{masys2012technical,Kohane2015TenMedicine}.However, clinical decision support systems that assist physicians in interpreting complex patient data typically operate on a per-patient basis. This strategy does not exploit the extensive latent medical knowledge from the whole database and multiple data types. The emergence of large digital systems offers the opportunity to integrate population information actively into these tools. Patient similarity analysis is one way to utilize recently available digital data sets and facilitates integration from the whole database and multiple data types for data-driven medical decision making. 

\begin{figure}[h!]
  
  \label{flowchart}
  \centering
  \includegraphics[width=1\textwidth]{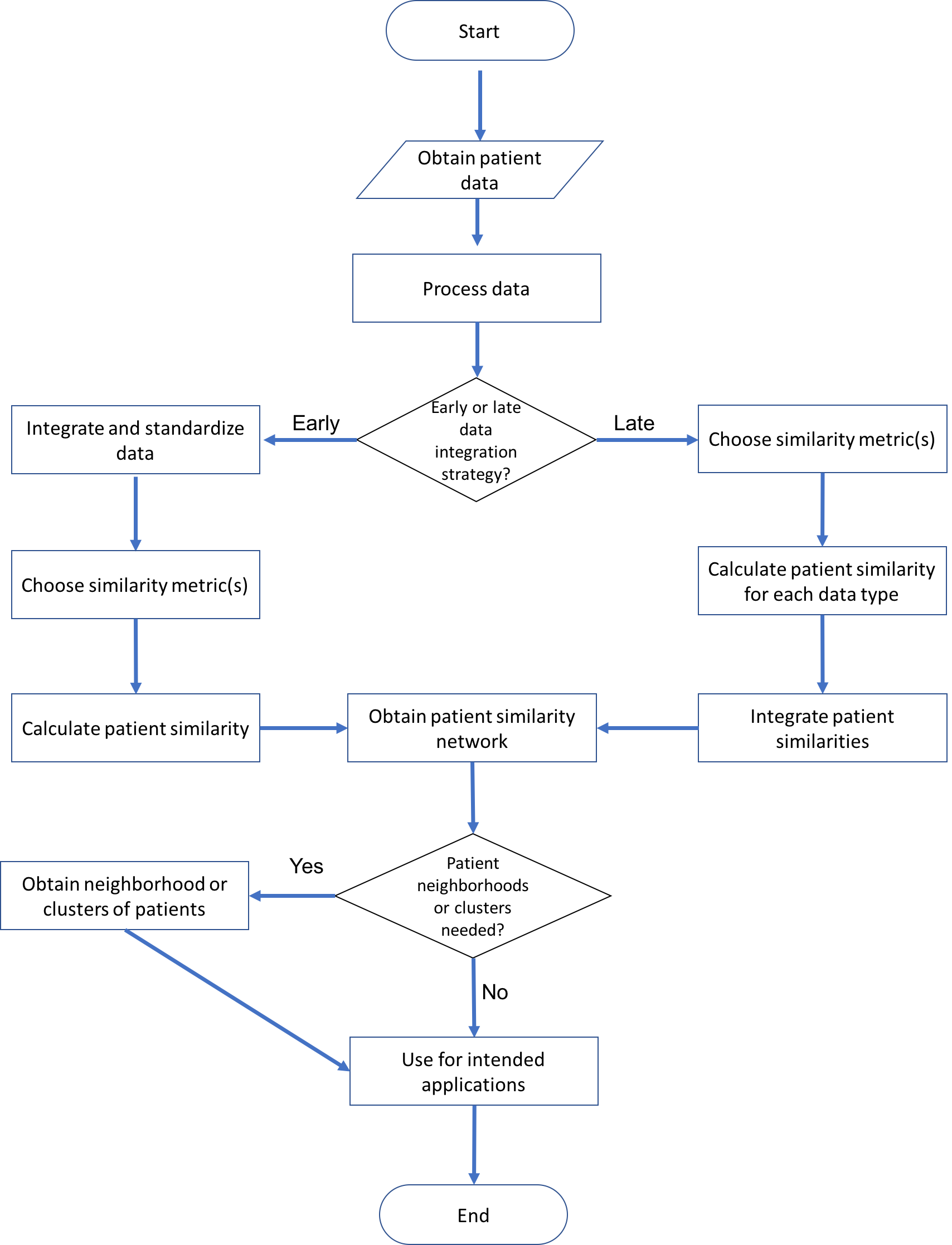}
  \caption{Flowchart of a typical patient similarity analysis.After obtaining patient similarity data, a patient similarity study usually starts with data processing and integration. A data integration strategy needs to be decided. Common options are early and late integration. Sometimes an intermediate integration strategy can be used(not shown in the flowchart, see Section \ref{Data integration strategies} for details)\cite{Gligorijevic2015MethodsChallenges}.Next, a patient similarity metrics (or a group of metrics) need to be chosen to quantitatively measure similarity score or distances between each pair of patients.Before the patient similarity network obtained being used for next stage of study, neighborhoods or clusters of patients are usually defined.}
\end{figure}

The patient similarity could take both phenotypic and genotypic data into account including but not limited to the electronic medical record, social media data, genomics, transcriptomics, proteomics, microbiomics, and other “omics”. When it comes to technical details, patient data types are often heterogeneous which means data types can exist in different forms such as continuous numerical, categorical, binary, and temporal. Patient similarity analytics calculate the similarity between each pair of patients based on these heterogeneous data. Therefore, one of the first steps to consider inpatient similarity analysis is data integration. After the data integration strategy is decided, the next step is to define a patient similarity metric so that distance or similarity scores between patients can be calculated in a systematic and consistent manner. Many analyses can be done after a patient similarity network is built but one common next step is to cluster patient-based similarity scores or define a group of patients that are most similar to the target patient (Figure \ref{flowchart}). In this review, applications of patient similarity and each of the technical steps mentioned above is explored. 

\clearpage

\section{Values and applications of patient similarity networks}
\label{Values and applications of patient similarity networks}

The effective utilization of different data about patients is important for medical activities such as physician decision support, comparative effectiveness research, disease modeling, and policymaking. Patient similarity measurements could possibly make contributions to these activities. 

\subsection{Patient and disease sub-typing }

Patient similarity has been used to identify sub-types of diseases. In the case of glioblastoma multiforme, patient similarity networks have been used to identify patient sub-groups, including a patient subgroup with a substantially more favorable prognosis and patients with favorable responses to the drug temozolomide\cite{Wang2014SimilarityScale}. The patient similarity was also used to obtain distinct subtypes of type 2 diabetes, each of which is associated with different diseases and specific genetic mutations. Similar sub-typing or grouping approaches have been applied to many other diseases like lung pneumonitis.\cite{Li2015IdentificationAccess,tirunagari2015identifying,Kohonen1982Self-organizedMaps,Chen2008UsingMap}.      

\subsection{Personalized medical predictive models}

Medical prediction is one of the key applications of big data, artificial intelligence, and other modern technologies in health care. Instead of one-fits-all approaches, medical outcome prediction and decision making based on personalized models could be more accurate, especially when the amount of data available is limited
\cite{liu2017deepfacelift,Lee2015PersonalizedMetric}. Patient similarity analysis would facilitate the development of personalized medical prediction. One such example is mortality prediction in intensive care units(ICUs) where patient similarity has been to boost the power of the model by using only patients most similar to the target patient in model training \cite{Lee2015PersonalizedMetric}. Also, a patient similarity network has been used to develop an automated method to infer an individual patient's discharge diagnosis based on electronic medical record \cite{Gottlieb2013ASimilarities}.

It has been suggested that network-based prediction which takes advantage of using the whole network of a patient sometimes has better performance than just using a subgroup of patients \cite{Wang2014SimilarityScale}.Therefore, it is often worth the time and effort to conduct a systematic patient similarity analysis to obtain the whole network before designing the model training strategy next step. 

\subsection{Personalized treatment design}

Patient similarity has also been used to design personalized treatment. Using techniques such as label propagation, the effectiveness of different treatments can be estimated for each patient using information from patient similarity network, drug-drug similarity network, and patient-drug prior association. This strategy has been used to study hypercholesterolemia patients' responses to different treatments \cite{Zhang2014TowardsAnalytics.}.

 In addition to the three types of networks mentioned above, many other data types could be utilized for personalized treatment design. These include but are not limited to disease-disease networks \cite{Kwang-Il2007TheNetwork},disease-symptom network \cite{Zhou2014HumanNetwork}, disease-gene \cite{Kwang-Il2007TheNetwork} and gene regulatory networks\cite{Liu2017UniversalNetworks,Albergante2016InsightsFoundations}.  

\subsection{Select informative patients}
Sometimes only a small portion of medical information is available for patients in a database or cohort of interest. It is often challenging to collect complete data from all patients given limited time and resources. With an established patient similarity network in hand, it is possible to decide quantitatively which patients are most informative to make a medical decision or build a predictive model. In research on breast and diabetes,  patient similarity networks derived from electronic medical records have been used to select potentially more informative patients from the population for experts to give feedback. Similar methods have been applied to analyze Alzheimer's disease using patient similarity estimated from MRI brain scan. \cite{Qian2015APrediction}.
\begin{figure}[ht!]
  
  \label{dataintegration}
  \centering
  \includegraphics[width=0.8\textwidth]{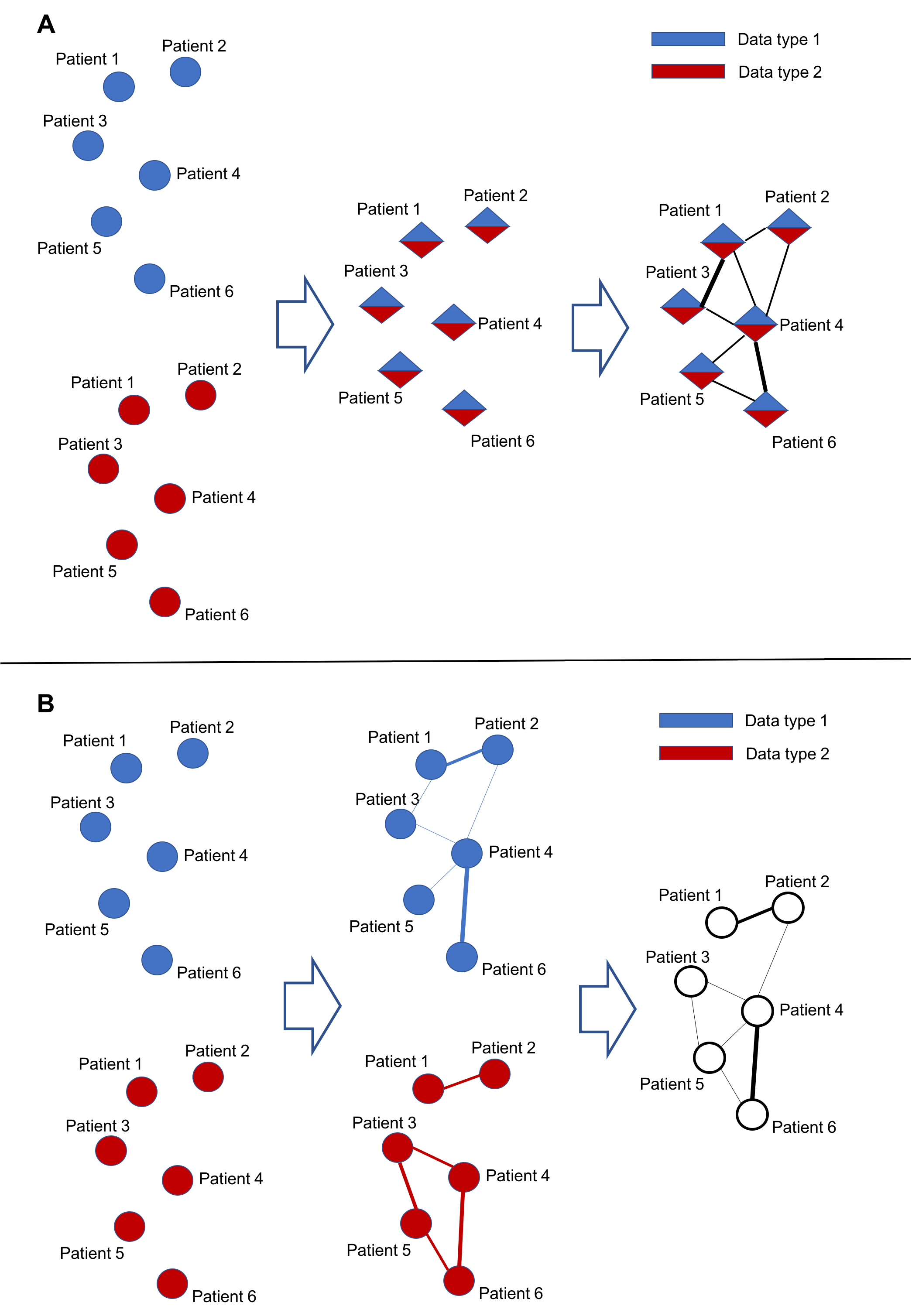}
  \caption{Early and late data integration strategy. A. In early data integration strategy, different data types are converted and standardized into the same format before measuring pairwise patient similarities. B. In late data integration strategy, a patient similarity network is built for each data type before the networks are merged into a single network. }
\end{figure}

\section{Data integration strategies}
\label{Data integration strategies}

The massive clinical and biological data around patients are highly heterogeneous. This means the data exist in different forms such as numeric values, categorical values, temporal waves, medical images. Although there are many similarity learning algorithms, they typically work with a single type of data.  The first step in patient similarity analysis is often to process the raw patient data so that they can be put into desired models. Therefore, one thing to consider before applying any sophisticated algorithm is how to integrate all the data of each patient. In the patient similarity analysis process, data integration can be generally categorized as early or late integration\cite{Gligorijevic2015MethodsChallenges}. 

Early data integration strategy combines different patient features into a single dataset, usually feature vector,  on which the data model is built (Figure \ref{dataintegration} A). This often requires a transformation of data types into a common representation. This transformation process may result in information loss. Late data integration builds a model for each data type (patient feature) and then combine them into a unified model(Figure \ref{dataintegration} B). Building models from each dataset in isolation from others disregard their mutual relations, often resulting in reduced performance of the final model

In early integration strategy, many methods represent each patient as a feature vector \(x_i \in R^p\) where \(i\) refers patient \(i\) and \(p\) is the number of features. If \(x_i\) is a vector of only numeric values, all patient data need to be converted into the same format such as a real number. For example, max, min, and the mean of a temporal measurement can be used. In an early feature integration strategy, A similarity metric  \(S(x_i,x_j)\)is used to measure the similarity between patients (or distance metric \(D\) can bed used which could be roughly understood as \(\frac{1}{Similarity}\)). Further analysis can be conducted based on \(S(x_i,x_j)\). There are many possible options for the metric \(S(x_i,x_j)\). The choice should be informed by the problem of interest and format of input data (See details in Section \ref{Similarity/distance metrics}). 

In late integration strategy , instead of obtaining a distance \(S(x_i,x_j)\) between \(x_i\) and \(x_i\), a list of similarity scores or distances \(S_1(x_i,x_j),S_2(x_i,x_j)...S_p(x_i,x_j)\) are calculated for each data type. All the \(p\) similarity scores or distances between \(x_i\) and \(x_i\) are then integrated into a a single score. Features do not need to be converted into the same format and different similarity/distance metrics can be used between for each data type. For examples,  \(S_1(x_i,x_j)\) can be the euclidean distances between numeric measurement and \(S_2(x_i,x_j)\) can be the correlation between two temporal measurements. There are many ways to combine multiple similarity measures.

There is also another strategy called intermediate integration which combines data through inferring a joint model. This strategy does not require any data transformation and thus, it theoretically does not result in any information loss\cite{Gligorijevic2015MethodsChallenges,liu2019integrative}. However,to the best of our knowledge, intermediate data integration has not been widely used in patient similarity analysis yet.

\section{Similarity/distance metrics}
\label{Similarity/distance metrics}
There is more than one way to quantitatively measure similarity or distance between each pair of patients. The methods can be generally classified into unsupervised, supervised, and semi-supervised. Unsupervised methods obtain similarity among patients directly using predefined static similarity or distance metrics to unveil hidden patterns in the datasets.  In supervised approaches, "real similarity" labels are given to pairs of patients which are usually obtained from clinician experts and medical diagnoses. The patient similarity metrics are calculated according to the labels. Semi-supervised methods fall between unsupervised and supervised methods and utilized both labeled and unlabeled data. \cite{jia2019using,kato2020clustering}

\subsection{Unsupervised methods: static similarity/distance metrics }

In early integration strategy , there are also many potential methods to measure similarity or distances between feature vectors of each pair of patient vectors(\(x_{1},x_{2}....x_{n} \in R^p\) and \(p>1\)). One of the simplest options is Euclidean distance which measures straight line distances between points in Euclidean space. Standardization is necessary if scales of features differ.   
\[D(x_i,x_j)=\sqrt{(x_i-x_j)^T (x_i-x_j)}{}=\sqrt{\Sigma _{n=1} ^{p} (x_{in}-x_{jn})^2}{}\]\cite{Ng2015PersonalizedSimilarity}.

One widely used distance metric in patient similarity study is Mahalanobis distance\cite{Lowsky2013AMethod,P.C.Mahalanobis1936OnStatistics} or variations of it\cite{panahiazar2015using}. The equation of  Mahalanobis distance is as follow:
\[D(x_i,x_j)=\sqrt{(x_i-x_j)^T \Sigma _S ^{-1} (x_i-x_j)}{}\]
where \(\Sigma\) is the the covariance matrix for all \(x\). Mahalanobis distance is a generalized method to measure how many standard deviations exist between two points or a a point and a distribution. Mahalnoibis distance is unit less and scale-invariant which means it is independent of the units in which the covariants are expressed. When \(\Sigma\) is an identity matrix , it is equivalent to Euclidean distance. 

Another similarity metric frequently used in patient similarity measurement is cosine similarity\cite{Li2015IdentificationAccess,Lee2015PersonalizedMetric,Lee2016PersonalizedBagging}. \[S(x_i,x_j)=\frac{x_i\cdot x_j}{\parallel x_i\parallel _2  \parallel x_j\parallel _2}=\frac{\Sigma _{n=1} ^{p} x_{in} \times \Sigma _{n=1} ^{p} x_{jn}}{\sqrt{\Sigma _{n=1} ^{p} x_{in} ^2} \sqrt{\Sigma _{n=1} ^{p} x_{jn} ^2} } \]
Since the patient similarity metric is an angle cosine, it normalizes between −1 (meaning exactly opposite) and 1 (meaning exactly the same).

Jaccard similarity coefficient is a measurement of similarity between finite sample sets and is defined as the size of intersection divided by the size of the union. Jaccard similarity was used to estimate similarity between patient vectors with binary attributes \cite{Zhang2014TowardsAnalytics.}.

\[S(x_i,x_j)=\frac{|x_i \cap x_j|}{|x_i \cup x_j|}\]

Where \(S(x_i,x_j)\) is the Jaccard similarity coefficient. Patient vectors \(x_i\) and \(x_j\) are of the same length \(p\) and with binary inputs \(0\) or \(1\). \(x_i \cap x_j\) is the intersection between \(x_i\) and \(x_j\),meaning attributes of the same values in the both vector.\(x_i \cup x_j\) is the union of the two vectors,meaning all the \(p\)attributes in each vector.\(|\dot|\) is length of the vector.

In late integration strategy, similarity/distances scores for each data types are usually calculated separately and the choice of metrics are highly dependent on the type of data. For example, for scalar data, \(x_{i},x_{j} \in R^p\),where \(p=1\). \(S(x_i,x_j)\) can be considered as \(\frac{1}{1+|x_i-x_j|}\)\cite{Chan2010MachineChemotherapy}. For binary data type, the similarity score can be calculated as whether \(x_i\) and \(x_j\) have the same value,ie \(S(x_i,x_j)=1-\frac{|x_i-x_j|}{|x_i|}\)\cite{Qian2015APrediction}. These obtained independent similarity measures can then be integrated different similarity networks in unsupervised manner using methods like network fusion or matrix factorization \cite{Wang2014SimilarityScale,liu2019integrative,Zitnik2015DataFactorization,Zitnik2015DataFactorization}.

\subsection{Supervised patient similarity metrics }
Sometimes, patient similarity analysis is highly context sensitive and depends depends on factors such as diseases and stages of diseases. An alternative to calculating patient similarity using static metrics is using metrics obtained from supervised learning. This means estimating the patient similarity according to certain ground truth label. 

One such approach is to use physician's judgment of patient similarity as gold standard.  A distance metric can be designed to automatically adjust the importance of each features according to physician's belief. In early integration strategy, one such option is to learn a generalized Mahalanobis distance which is defined as the equation below \cite{Sun2010LocalizedData,Sun2012SupervisedRecords} :

\[d_gm(x_i,x_j)=\sqrt{(x_i-x_j)^T P (x_i-x_j)}{}\]

Where \(P\) is a \(N\) by \(N\) precision matrix. \(N\) is the number of patients. Matrix \(P\) is positive semi-definite and is used to incorporate correlations between different feature dimensions. The goal is to learn the optimal \(P\) such that patients with the same label from physician's feedback are close while the patient with different labels is away from each other. The optimal \(P\) can be obtained using approaches such as decomposed Newtown's method \cite{Jia2009TraceRevisited,li2020self}.

 A similar approach has been used to get metrics trained using disease onset as a label. The metric obtained from this supervised learning has been shown to perform better than static metrics when used for diabetes onset prediction and risk factor profiling\cite{Ng2015PersonalizedSimilarity,Wang2009TwoClassification}.    When using a late integration strategy to combine multiple similarity measures, one simple way is to assign a weight to each measure.
\[S(x_i,x_j)=\Sigma_{n=1}^{p} \sigma _n S_n(x_{in},x_{jn})\] Where \(S_n\) is the similarity measure for the \(n^{th}\) data type between each pair of patients(\(x_{in}\) and \(x_{jn}\)). \(\sigma _n\) is the weight for the corresponding similarity measure and its value can be learned in a supervised manner using methods such as logistic regression and Cox regression \cite{Klenk2010DeterminingNetworks}.

\subsection{Semi-supervised method}
When doing medical data analysis, it is often that physician's feedbacks or diagnosis for certain diseases is only available for some patients. This makes some of the data labeled while others are unlabeled. In this scenario, the patient similarity metric can be obtained using a semi-supervised approach.
A practical method to do design a low dimensional space in which the distances between each pair of the patient are transformed into. An optimal space is identified by minimizing a cost function considering both labeled and unlabeled data. Let the original patient feature vector be \(x_1,x_2...x_n \in R^p\).\(F\) is the set of patient vectors transformed into a lower-dimensional space with \(F={f_1,f_2...f_n}\) and \(f_i in R^q\), usually \(q<p\)\cite{Wang2015PSF:Supervision,wang2012medical,Zhou2017ALearning,Wu2010PredictionData} . Then the objective function is: \[J_{min F}(F)=U(F)+\lambda L(F)\] \(F\) is designed to minimize \(J(F)\) such that the distance between unlabeled patient vector with similar feature values (\(U(F)\)) are small and distance between labeled pairs match the ground truth (\(L(F)\)). Methods such as PCA or local spine regression can be used to capture \(U(F)\) and weighted Euclidean distance can be used for\(L(F)\). The exact optimization algorithms can be found in the original publications \cite{Wang2015PSF:Supervision,wang2012medical}  \subsection{Neural Network Learning} Neural Network Learning more and more popular in recent years. Neural Network is one of the machine learning algorithms to analyze image voice and some data-set contains the heavy size of data. It is a network structure composed of many simple elements. This network structure is similar to the biological nervous system and is used to simulate the interaction between living things and the natural environment. In CNN(Convolutional Neural Network), analyze Electronic Health Records will provide the similarity between all patient pairs and it can predict the health status of patients.\cite{zeng2019measure,lin2020medical}.
\subsection{Active learning} The key idea of active learning is that a machine learning algorithm can achieve higher accuracy with fewer training labels if it is allowed to choose the data from which it learns. Active learning extends machine learning by allowing learning algorithms to typically query the labels from an oracle for currently unlabeled instances \cite{Qian2015APrediction,suo2018deep}. To obtain similarity metrics using supervised or semi-supervised strategies, one challenge is that getting labeled data, eg. similarity estimation from experts, in medical settings is usually expensive and time-consuming. Using an active learning strategy, such as counting set cover, to select the most informative patient pairs or groups from the cohort would make studies efficient \cite{Qian2015APrediction,rahman2020phenotyping,zhan2016low}.

\begin{table}
\caption{Representative methods for and applications of patient similarity analysis}
\label{Methods}
\begin{adjustbox}{max width=\textwidth,totalheight=\textheight-2\baselineskip,center}
\begin{tabular}
{m{0.16\textwidth}|m{0.15\textwidth}|m{0.15\textwidth}|m{0.1\textwidth}|m{2.5cm}|m{2cm}}
    \hline
    Similarity metrics & metric type & neighborhood/ cluster detection & data integration strategy & Application & Publication\\
    
    \hline
    Euclidean distance&Unsupervised  &spectral clustering  & late &disease sub-typing , survival risk prediction &Wang\textit{et al.} 2014 \cite{Wang2014SimilarityScale}\\
    \hline
     Euclidean distance&Unsupervised &SOM & early&self-caring behaviour study for diabettes&Tirunagari \textit{et al.} 2015  \cite{tirunagari2015identifying}\\
     \hline
    Euclidean distance&Unsupervised &SOM & early &lung radiation pneumonitis risk prediction & Chen \textit{et al.} 2008  \cite{Chen2008UsingMap}\\
     \hline
     Euclidean distance&Unsupervised& SOM /Gaussian mixture model& early&hospital length of stay prediction& Gorunescu\textit{et al.} 2010  \cite{Gorunescu2010PatientSystem}\\
     \hline
     Euclidean distance in transformed space&Semi-supervised &KNN &early&heart failure prediction &Wang \textit{et al.}2015  \cite{Wang2015PSF:Supervision}\\
    
    \hline
    Cosine similarity &Unsupervised  &KNN &early  &post discharge mortality prediction &Lee \textit{et al.} 2015 \cite{Lee2015PersonalizedMetric}\\
    \hline
    Cosine similarity&Unsupervised  &Topology based clustering  &early &subgroups identification of type 2 daibetes& Li \textit{et al. 2015} \cite{Li2015IdentificationAccess}\\
    \hline
    Cosine similarity &Unsupervised  & Weighted sampling &early  &post discharge mortality prediction&Lee \textit{et al.} 2016 \cite{Lee2016PersonalizedBagging}\\
    \hline
    Tanimoto coefficient & Unsupervised & N.A.& early &personalized treatments for hypercholesterolemia&Zhang \textit{et al.}2014 \cite{Zhang2014TowardsAnalytics.}\\
    
    \hline
    Mahalanobis distance & Unsupervised  & KNN & early &predict survival curve of kidney transplantation & Lowsky \textit{et al.} 2013 \cite{Lowsky2013AMethod}\\
    \hline
    Mahalanobis distance&Unsupervised &K-means/ hierarchical clustering/supervised classification&early&hear failure therapy recommendation&Panahiazar \textit{et al.} 2015 \cite{panahiazar2015using}\\
    \hline
    low rank Mahalanobis distance&Unsupervised/ Semi-supervised &KNN &early  &congestive heart failure prediction&Wang \textit{et al.}2012 \cite{Wang2012CompositeAssessment}\\
    
    \hline
    generalized Mahananobis distance &Supervised &KNN & early &classify arterial hypotensive episode&Sun\textit{et al.} 2010 \cite{Sun2010LocalizedData}\\
    \hline
    generalized Mahananobis distance &Supervised &KNN , K-means& early &combine feedbacks into patient similarity metrics &Sun \textit{et al.} 2012 \cite{Sun2012SupervisedRecords}\\
    \hline
     generalized Mahananobis distance&Supervised &KNN  &early &diabetes onset prediction and risk profiling&Ng \textit{et al.} 2015  \cite{Ng2015PersonalizedSimilarity}\\
    \hline
    Sharing diagnoses& Semi-supervised&fixed threshold  &N.A.&risk prediction of multiple diseases&Qian\textit{et al.}2014 \cite{Qian2015APrediction}\\
    
    \hline
    multiple similarity metrics&Unsupervised  &N.A. & late &inferring discharge diagnosis&Gottlieb \textit{et al.}2013 \cite{Gottlieb2013ASimilarities}\\
    \hline
     multiple similarity metrics& Unsupervised  &N.A &late&treatment recommendation &Wang \textit{et al.}2015  \cite{Wang2015AnSimilarity}\\
    \hline
     multiple similarity metrics&Supervised &Supervised classification &late  & cancer chemotherapy survival prediction &Chan\textit{et al.}2010 \cite{Chan2010MachineChemotherapy}\\
     
    \hline
     multiple similarity metrics&Supervised &N.A. &late  & Medical social medium recommendation  &Klenk\textit{et al.}2010 \cite{Klenk2010DeterminingNetworks}\\
    \hline
     patient-similarity-based model&supervised & KNN& late&medical recommendation &Zheng\textit{et al.} \cite{jia2019using}\\ 
    \hline
    Multi-modal Data for Retinal Disease Diagnosis& self-supervised&NA & early&medical and treatment recommendation &Li\textit{et al.}\cite{li2020self} \\
    \hline
    
    \hline
    GAN model for clinc treatment &self-supervised&GAN&early& medical and treatment advice &Yang\textit{e  al.}\cite{yang2019gan}\\
    \hline
    Risk-based forecasting & supervised &NA & late & forecasting advice & Oviedo\textit{et  al.}\cite{oviedo2019risk}\\
    \hline
    lung cancer classification & supervised & NA & late & medical advices & Ahmed\textit{et al.}\cite{ahmed2019lung}\\
    \hline
\end{tabular}
\end{adjustbox}
\end{table}

\section{Patient neighborhoods and clusters detection methods}
\label{Patient communities and clusters detection methods}
With patient similarity network(s) in place, one common practice is to define the neighborhood of patients who are most similar to the patient of interest or divide the population into clusters.

\subsection{Neighborhood }
One way to define a neighborhood is to set a fixed threshold of similarity score or distance. If a patient's similarity score is larger than the threshold or distance is smaller than the threshold, he/she is considered a neighbor to the patient of interest \cite{Qian2015APrediction}. 

Another unsupervised approach is is K-nearest neighbors(KNN).  K-nearest neighbors to \(x_i\),\(S_i^K \in S \), are chosen. Choice of \(K\) can be chosen using heuristic methods such as separating \(S\) into training and validation sets\cite{Lowsky2013AMethod,wang2012medical,Lee2015PersonalizedMetric,Sun2010LocalizedData,Ng2015PersonalizedSimilarity,jia2020patient}.

A weighted sampling of similar patients has also been used to get a neighborhood of similar patients. Firstly, pairwise similarity values between every patient and the patient of interest are calculated. A fixed number of patients are sampled from the population in a probabilistic manner using the similarity values as weights \cite{Lee2015PersonalizedMetric}. The sampled patients are then used for further study. This sampling process can repeat multiple times. If each sample was used to train a separate predictive model.  The approach can be considered as a form of ensemble learning. 

\subsection{Clusters}
Sometimes, instead of identifying neighbors of the patient of interest, it is more useful to organize the patient population into clusters(or groups). 
 
K-means algorithm is one of the most popular algorithms in current use due to its simplicity and computational speed \cite{raykov2016k}.For a set of patient vectors \(x_1 ,x_2 ....,x_n \in R^p\),K-means clustering aims to partition the n observations into K(K<=n) sets \((S_1,S_2....S_k)\). This object is to minimize within cluster sum of squares:

\[ Argmin_S \Sigma _{i=1} ^{k} \Sigma _{x \in S_i} \parallel x-u_i \parallel \]

The algorithm to obtain K-means clusters usually includes selecting K, randomly choosing initial means, associate each patient to means, recalculating the new centroid, and repeat until converge. As K-means is a heuristic algorithm, there is no guarantee that it will converge to a global minimum.

 Another method used for patient clustering is spectral clustering  \cite{Wang2014SimilarityScale}. Spectral clustering uses the spectrum (eigenvalues) of the similarity matrix of the data to perform dimensionality reduction before clustering in fewer dimensions. Let \(W\) be the symmetric similarity matrix , where \(A_{ij}\) is the similarity between patient \(i\) and patient \(j\) (i.e between \(x_i\) and \(x_j\)).  A general approach is to use a standard clustering method (eg. K-means) on relevant eigenvectors of a Laplacian matrix of \(A_{ij}\). 

Maximuma-posteriori Dirichlet process mixtures is a method developed recently and has applied to patient clustering \cite{raykov2016k}. It is related to nonparametric Bayesian Dirichlet process mixture modeling and has a couple of advantages such as being able to automatically estimate number of clusters from the data.

Self-organizing map(SOM) is a type of artificial neural network that is trained in a unsupervised manner to conduct non-linear transformation of the input space into low dimension. Different from many other artificial neural networks, SOM uses competitive learning as opposed to error-correction learning (such as backpropagation with gradient descent). SOM uses a neighborhood function to preserve the topological features of the input space. There are two stages in using SOM , training and mapping. In training stage, the low dimensional representation (map) is built. In mapping stage, a new input is classified. A SOM consist of many neurons.Each neuron has a weight vector. The weight vector of neuron \(t\) is \(\omega_t \in R^p\). \(t=1,2,3....m\).\(m\) is the number of output neurons.\(p\) is the dimension of input vector. An input vector (\(x_i \in R^p\)) is mapped by finding the the neuron with closest weight vector which can be written as 
\[arg_{min_t} D(x_i,\omega_t)\]
Where D is the distance measure between vectors such as Euclidean distance or Mahalanobis distance.SOM with a small number of neuron behaves similar to K-means.SOM has been used to visualize and understand to obtain similar patients clusters for diabetes , lung pneumonitis and other diseases.\cite{tirunagari2015identifying,Kohonen1982Self-organizedMaps,Chen2008UsingMap}

\section{Discussion and future opportunities}

The patient similarity could be considered a branch of person similarity research which is a big topic. Compared with many other person similarity fields such as user preference, the amount of research that has been conducted on patient similarity is relatively limited\cite{Lee2015PersonalizedMetric,breese2013empirical,Herlocker2004EvaluatingSystems}.  There are many challenges in detecting patients' similarities such as the heterogeneous nature of medical data and inconsistency between different data sources. Fortunately, many of the roadblocks can be simplified and conquered using technologies adapted from other fields. At the data integration stage, intermediate data integration technology such as non-negative matrix factorization methods used in studying disease-disease association and human chromatin interaction can be adapted to minimize information loss \cite{Zitnik2013DiscoveringData,Zitnik2015DataFactorization,liu2019integrative}.More similarity metrics can be introduced into detecting patient similarity, especially supervised and semi-supervised methods.    At the neighborhood/cluster detection stage, many advanced technologies like growing SOM and semi-supervised clustering can be utilized\cite{kingma2014semi,Zhu2010TheCodes}. 
One important field to which patient similarity network could make a contribution to precision medicine which can be considered a special type of personalized data science. A variety of personalized data analytic has been used in fields like product recommendation and consumer credit \cite{adomavicius2005toward,Hand1997StatisticalReview}. Personalized data-driven clinical analysis is still a developing field. Recently, there are a couple of efforts such as applying personalized machine learning for pain prediction in medical settings \cite{liu2017deepfacelift}. With patient similarity networks in place, a lot more research can be done and many modern technologies such as domain adaptation and transfer learning can be applied to achieve high accuracy\cite{glorot2011domain}.

\bibliographystyle{alpha}
\bibliography{sample,Mendeley}

\end{document}